\pdfoutput=1
\documentclass[11pt]{article}

\usepackage{emnlp2021}

\usepackage{times}
\usepackage{wrapfig}
\usepackage{multirow}
\usepackage{graphicx}
\usepackage{amsmath}
\usepackage{latexsym}
\usepackage{hyperref}
\usepackage{array}
\usepackage[T1]{fontenc}
\usepackage[utf8]{inputenc}

\usepackage{microtype}

\title{Privacy enabled Financial Text Classification using\\ Differential Privacy and Federated Learning}

\author{Priyam Basu \and Tiasa Singha Roy \\
  Manipal Institute of Technology \\
  \texttt{\{priyam.basu1, tiasa.singharoy\}@learner.manipal.edu} \\
  
  \AND
  Rakshit Naidu \\
  Carnegie Mellon University \\
  \texttt{rnemakal@andrew.cmu.edu} \\\And
  Zumrut Muftuoglu \\
  Yildiz Technical University\\
  \texttt{zumrutmuftuoglu@gmail.com}}

\begin{document}
\maketitle
\begin{abstract}
Privacy is important considering the financial Domain as such data is highly confidential and sensitive. Natural Language Processing (NLP) techniques can be applied for text classification and entity detection purposes in financial domains such as customer feedback sentiment analysis, invoice entity detection, categorisation of financial documents by type etc. Due to the sensitive nature of such data, privacy measures need to be taken for handling and training large models with such data. In this work, we propose a contextualized transformer (BERT and RoBERTa) based text classification model integrated with privacy features such as Differential Privacy (DP) and Federated Learning (FL). We present how to privately train NLP models and desirable privacy-utility tradeoffs and evaluate them on the Financial Phrase Bank dataset.
\end{abstract}

\section{Introduction}
% Natural language processing (NLP) is one of sub-topic of artificial intelligence that provides interaction between humans and computers by understanding and processing human languages (??). It tackles computational linguistics with statistics,machine learning and deep learning(??). Machine translation, speech recognition, sentiment analysis,text classification are examples for the main NLP application we could see in real life.

Divulging personally identifiable information during a business transaction has become a commonplace occurrence for most individuals. This activity can span from sharing of bank account numbers, loan account numbers, and credit/debit card numbers, to providing non-financial personally identifiable information such as name, social security number, driver’s license number, address, and e-mail address. Maintaining the privacy of confidential customer information has become essential for any firm which collects or stores personally identifiable data. The financial services industry operates and deals with a significant amount of confidential client and customer data for daily business transactions. Though many organizations are taking strides to improve their privacy practices, and consumers are
becoming more privacy-aware, it remains a tremendous burden for users to manage their privacy~\cite{1281243}.

NLP has major applications in the finance industry for many tasks such as detection of entities for gross tax calculation from invoice and payroll data, categorising different kinds of financial documents based on type, grouping of financial documents based on semantic similarity, sentiment analysis of financial text~\cite{vicari2020analysis}, conversational bots for banking systems, investment recommendation engines etc. 

Text Classification can be extended to many NLP applications including sentiment analysis, question answering, and topic labeling . For example, financial or government institutions that wish to train a chatbot for their clients cannot be allowed to upload all text data from the client-side to their central server due to strict privacy protection statements~\cite{liu2021federated}. At this point, applying the federated learning paradigm presents an approach to solve the dilemma due to its advances in privacy preservation and collaborative training where the central server can train a powerful model with different local labeled data at client devices without uploading the raw data considering increasing privacy concerns in public.

The goal of this paper is to propose a privacy enabled text classification system, combining state-of-the-art transformers (BERT and RoBERTa) with differential privacy, on both centralized and FL based setups, exploring different privacy budgets to investigate the privacy-utility trade-off and see how they perform when trying to classify financial document-based text sequences. For the federated setups, we try to explore both IID (Independent and Identically Distributed) and non-IID distributions of data.
% \begin{figure*}[h]
% \vskip 0.0in
% \begin{center}
% \centerline{\includegraphics[width=12cm]{pipeline2.jpeg}}
% \caption{Pipeline of proposed model}
% \label{fig:Figure1}
% \end{center}
% \vskip -0.0in
% \end{figure*}

\begin{figure}
%\vskip 0.2in
%\begin{center}
\includegraphics[width=\linewidth]{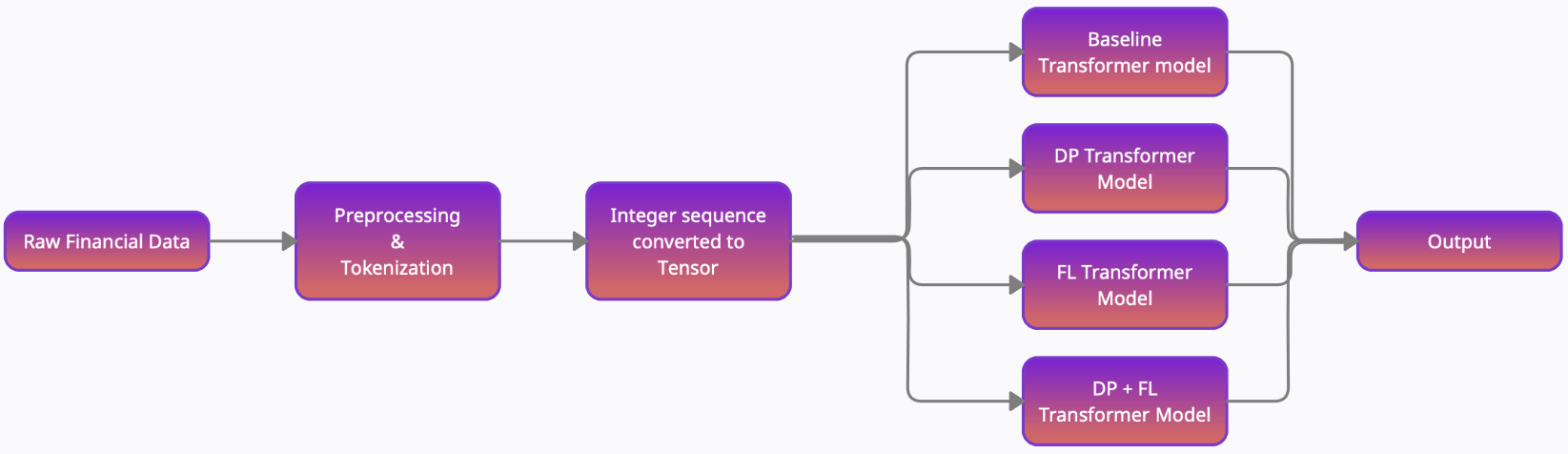}
% \vspace{-5ex}
\vspace{0.2in}
\caption{Pipeline}
\vspace{-0.2in}
% \vspace{-5ex}
\label{fig:Figure1}
%\end{center}
\end{figure}

% \begin{figure*}[h]
% \vskip 0.0in
% % \begin{center}
% \includegraphics[width=\linewidth]{pipeline2.jpeg}
% \caption{Pipeline of proposed model}
% \label{fig:Figure1}
% % \end{center}
% \vskip -0.0in
% \end{figure*}

\section{Related Work}
Deep learning  techniques have often been used to learn text representations via neural models by language application. The input text can give us individual demographic information about the author. Sentiment analysis can be used for the classification or categorization of financial documents.~\citeauthor{xing2020financial} investigate the error patterns of some widely acknowledged sentiment analysis methods in the finance domain.~\citeauthor{9142175} perform more than one hundred experiments using publicly available datasets, labeled by financial experts. In their work,~\citeauthor{liu2020finbert} propose a domain-specific language model pre-trained on large-scale financial corpora and evaluate it on the Financial Phrase Bank dataset.~\citeauthor{araci2019finbert} presents a BERT-based model which is pre-trained on a large amount of finance-based data in his study.

Studies have been conducted on training differentially private deep models with the formal differential privacy approach in the literature~\cite{Abadi_2016, mcmahan2018learning, Yu_2019}.~\citeauthor{de45936660fb44f4a9ae286ca45be95c} discuss the security through differential privacy in textual data.~\cite{panchal2020differential} in his work portrays the use of DP in the generation of contextually similar messages for Honey encryption which encrypts messages using low min-entropy keys such as passwords. Federated learning is another privacy-enhancing approach~\cite{mcmahan2017communicationefficient, yang2019federated, kairouz2021advances, jana2021investigation, priyanshu2021fedpandemic}, which relies on distributed training of models on devices and sharing of model gradients. 
% There have been studies focusing on coping with the statistical challenges ~\cite{smith2018federated, Zhao18} and security concerns~\cite{Segal17, geyer2018differentially}.
~\citeauthor{liu2021federated} show how FL can be used for decentralized training of heavy pre-trained NLP models.~\citeauthor{basu2021benchmarking} in their work have shown a detailed benchmark comparison of multiple BERT based models with DP and FL for depression detection.~\citeauthor{jana2021investigation} in their work show a differentially private sequence tagging system in a federated learning setup.

\section{Dataset}
The key arguments for the low utilization of statistical techniques in financial sentiment analysis have been the difficulty of implementation for practical applications and the lack of high- quality training data for building such models. Especially in the case of finance and economic texts, annotated collections are a scarce resource and many are reserved for proprietary use only. For this reason, we use the Financial Phrase Bank dataset~\cite{malo2014good} which was also used for benchmarking the pre-trained FinBERT model for sentiment analysis~\cite{araci2019finbert}. The dataset includes approximately 5000 phrases/sentences from financial news texts and company press releases. The objective of the phrase level annotation task is to classify each example sentence into a positive, negative or neutral category by considering only the information explicitly available in the given sentence. Since the study is focused only on financial and economic domains, the annotators were asked to consider the sentences from the viewpoint of an investor only; i.e. whether the news may have a positive, negative or neutral influence on the stock price. As a result, sentences that have a sentiment that is not relevant from an economic or financial perspective are considered neutral. 

Given a large number of overlapping annotations (5 to 8 annotations per sentence), there are several ways to define a majority vote-based gold standard. To provide an objective comparison, the authors have formed 4 alternative reference datasets based on the strength of majority agreement. For the purpose of this task, we use those sentences with 75\% or more agreement. The final dataset has 3453 sentences in total out of which 60\% belong to the neutral class, 28\% belong to the positive class and 12\% belong to the positive class.

\section{Preliminaries}
Today, the text is the most widely used communication instrument.For years, researchers are studies focusing on implementing different approaches that make possible machines to imitate human reading~\cite{ly2020survey}. Natural Language Processing(NLP) lays a bridge between computers and natural languages by helping machines to analyze human language~\cite{10.5555/311445}
.\citeauthor{devlin-etal-2019-bert} developed a model which is based on bidirectional encoder representation~\cite{alyafeai2020survey}. RoBERTa is a modified form of BERT~\cite{liu2019roberta}.

\subsection{BERT}
Transformer-based models have been used since they use a self-attention mechanism and process the entire input data at once instead of as a sequence to capture long-term dependencies for obtaining contextual meaning. Bidirectional Encoder Representations from Transformers (BERT)\cite{devlin2018bert} tokenizes words into sub-words (using WordPiece) which are then given as input to the model. It also uses positional embeddings to replace recurrence.

\subsection{RoBERTa}
Robustly Optimized BERT-Pretraining Approach (RoBERTa)~\cite{liu2019roberta} is a state-of-the-art transformer model which improves BERT~\cite{devlin2018bert} that uses a multi-headed attention mechanism which enables it to capture long term dependencies. It essentially fine-tunes the original BERT model along with data manipulation and uses Byte-Pair Encoding for utilizing the character and word level representations and removed Next Sentence Prediction (NSP) to match or even slightly improve downstream task performance.

\subsection{Differential Privacy}
Differential Privacy (DP) is a privacy standard which allows data use in any analysis by presenting mathematical guarantee  ~\cite{DworkRoth14}. It provides strong confidentiality in statistical databases and machine learning approaches through mathematical definition. This definition is an acceptable measure of privacy concern~\cite{Dwork08}.

\textbf{Definition 1.1 : }\textit{$M$ and $E$ denote a random mechanism and each event (output) respectively. $D$ and $D'$ are defined neighboring datasets having difference with one record. ($\varepsilon$, $\delta$) protects confidentiality~\cite{Dwork11}.\textit{} M gives ($\varepsilon$, $\delta$)-differential privacy for and D and D' if M satsifies:}
\begin{equation}
\Pr\left[M\left(D\right) \in E\right] \leq e^{\epsilon }\cdot \Pr\left[M\left(D'\right) \in E\right] + \delta
\end{equation}
where $\varepsilon$ denotes the privacy budget and $\delta$ represents the probability of error. 

\subsubsection{The Privacy Budget}
The privacy guarantee  level of $M$ is controlled through privacy budget of $\epsilon$~\cite{Haeberlen11}.There are two widely used privacy budget compositions as the sequential composition and the parallel composition.

The ratio between the two mechanisms ($M(D)$ and $M(D')$) limits by $e^\varepsilon$. For $\delta$ $=$ $0$, M gives $\varepsilon$-differential privacy by its strictest definition. In other case, for some low probability cases, ($\varepsilon$,$\delta$)-differential privacy provides latitude to invade strict $\varepsilon$-differential privacy. $\varepsilon$-differential privacy is called as pure differential privacy and ($\varepsilon$, $\delta$)-differential privacy, where $\delta$ $>$ $0$, is called as \textit{approximate differential privacy}~\cite{beimel2014private}. Differential privacy has two implementation settings: Centralized DP (CDP) via DP-SGD and Local DP (LDP)~\cite{qu2021privacyadaptive}.

In CDP, a trusted data curator answers queries or releases differentially private models by using randomisation algorithms~\cite{DworkRoth14}. In this article, we use DP-SGD (Differentially Private Stochastic Gradient Descent)~\cite{Abadi_2016} to train our models.

% \begin{table}[h!]
% \centering
%   \caption{Averaged Test Accuracies of FL and DPFL models}
%   \begin{tabular}{|c|c|c|c|}
%   \hline
%   \textbf{Setup}	&\textbf{Epsilon}	&\textbf{BERT}	&\textbf{RoBERTa}\\
%   \hline 
%     \multirow{5}{4.5em}{Centralized DP}	
%     &0.5 &{31.5 $\pm$ 23.94} \&{31.36 $\pm$ 26.35}\\
%     & 5	&{37.48 $\pm$ 20.42} &{38.34 $\pm$ 20.08}\\ 
%     & 15 &{51.71 $\pm$ 14.71} &{51.34 $\pm$ 15.45}\\ 
%     & 20 &{55.37 $\pm$ 5.49} &{55.54 $\pm$ 5.54}\\ 
%     & 25 &{60.03 $\pm$ 1.37} &{62.6 $\pm$ 4.24}\\ 
%     \hline
%     \multirow{5}{4.5em}{DP-FL\\ IID}	
%     &0.5 &{14.57 $\pm$ 2.86} &{20.11 $\pm$ 7.68}\\
%     & 5	&{30 $\pm$ 25.6} &{30.04 $\pm$ 28.22}\\ 
%     & 15 &{40.34 $\pm$ 20.55} &{50.26 $\pm$ 20.84}\\
%     & 20 &{51.05 $\pm$ 7.95} &{54.78 $\pm$ 2.99}\\ 
%     & 25 &{53.47 $\pm$ 6.48} &{61.38 $\pm$ 0.93}\\ 
%     \hline
%     \multirow{5}{4.5em}{DP-FL\\ Non IID}	
%     &0.5 &{19.82 $\pm$ 5.97} &{33.13 $\pm$ 25.41}\\ 
%     & 5 &{35.74 $\pm$ 21.48} &{36.51 $\pm$ 26.87}\\ 
%     & 15 &{45.87 $\pm$ 15.56} &{49.83 $\pm$ 20.6}\\
%     & 20 &{52.43 $\pm$ 4.08} &{53.36 $\pm$ 3.27}\\ 
%     & 25 &{58.96 $\pm$ 2.56} &{60.83 $\pm$ 0.53}\\ 
%     \hline
%   \end{tabular}
% \end{table}

\subsection{Federated Learning}

As conventional centralized learning systems require that all training data produced on different devices be uploaded to a server or cloud for training, it may give rise to serious privacy concerns ~\cite{appledp17}.
FL allows training an algorithm in a decentralized way ~\cite{mcmahan2017communicationefficient, McMahan2016FederatedLO}. It ensures multiple parties collectively train a machine learning model without exchanging the local data ~\cite{li2021survey}. To define mathematically, it is assumed that there are $N$ parties, and each party is showed with $T_i$, where $i$ $\in$ $[1,N]$. For the non-federated setting, each party uses its local data and depicted by $D_i$ to train a local model $M_i$ and send the local model parameters to the server. The predictive data is sent only the local model parameters to the FL server. Most centralized setups have just the IID assumption for train test data but in a federated learning based decentralized setup, non-IID poses the problem of high skewness of different devices due to different data distribution~\cite{liu2021federated}.

% Recent works on language modelling (LM in Federated NLP mainly target on solving a word-level LM problem in mobile industry. Most works  consider variants of LSTMs as the client model. Given the limited computation budget on each device, we expect the parameter space of a neural language model to be as small as possible without degrading model performance. For federated optimization, existing federated optimization algorithms differ in client model aggregation on the server-side. 
In federated language modeling, existing works~\cite{yang2018applied} use FedAvg as the federated optimization algorithm. In FedAvg, gradients that are computed locally over a large population of clients are aggregated by the server to build a novel global model. Every client is trained by locally stored data and computes the average gradient with the current global model via one or more steps of SGD. 

Applying FL to text classification can cause problems such as designing proper aggregating algorithms for handling the gradients or weights uploaded by different client models.~\citeauthor{zhu-etal-2020-empirical} proposed a text classification using the standard FedAvg algorithm to update the model parameter with local trained models. Model compression has also been introduced to federated classification tasks due to the dilemma of computation restraints on the client-side, where an attempt to reduce the model size on the client-side to enable the real application of federated learning was made. For overcoming the communication dilemma of FL,  central server can successfully train the central model with only one or a few rounds of communication under poor communication scenarios in a one-shot or few-shot setting.

% \begin{figure}
% %\vskip 0.2in
% \begin{center}
% \includegraphics[width=\linewidth]{pipeline.jpeg}

% \vspace{-5ex}
% \caption{Pipeline of our benchmarking framework.
% }
% \vspace{-5ex}
% \label{fig:Figure1}
% \end{center}
% \end{figure}

\section{Experimental Results}

% \begin{table}[h]
%   \centering
%   \caption{Test Accuracies of Baseline Models}
%   \begin{tabular}{|c|c|c|c|c|}
%   \hline
%   \textbf{Baseline Model}	&\textbf{Accuracy}\\
%   \hline 
%     BERT	\text{67.71}\\ 
%     \hline
%     RoBERTa	\text{68.37}\\ 
%     \hline
%   \end{tabular}
%   \label{fin_dataset}
% \end{table}

In the scope of the study, the FinBERT (pre-trained) model is used as the base model. Two NLP models were trained by implementing DP and FL. In this section, the results presented in the tables are discussed. The results placed in the tables are the average and the standard deviation of the results obtained after running the models thrice.

The dataset was split into train set and test set with 80:20 train test ratio. BERT and RoBERTa based models were used for the language modelling part. It should be noted that the table contain the average and the standard deviation of the results obtained after running the models 3 times. Table 1 shows a comparison according to epsilon values between both the language models using Centralized DP and in a Federated Learning set up. The Opacus library was used along with PyTorch for the experiments. We implement DP, FL and DP-FL on BERT and RoBERTa for $\epsilon=0.5,5,15,20,25$. Our baseline model (with no noise) achieves an accuracy of \textbf{67.71\%} and \textbf{68.37\%} on BERT and RoBERTa respectively. 

In baseline mode, we can see that RoBERTa has a slight improvement over BERT because of its robustness owing to a heavier pre-training procedure.  We also notice that with the increase in epsilon values, the amount of standard deviation decreases as the model approaches towards its vanilla variant (without DP noise).

% \begin{table}[h]
%   \centering
%   \caption{Test Accuracies of Baseline Models}
%   \begin{tabular}{|c|c|c|c|c|}
%   \hline
%   \textbf{Baseline Model}	\textbf{Accuracy}\\
%   \hline 
%     BERT	\text{67.71}\\ 
%     \hline
%     RoBERTa	\text{68.37}\\ 
%     \hline
%   \end{tabular}
%   \label{fin_dataset}
% \end{table}

Table 2 also shows us the results obtained when DP was applied in a federated learning mode, both in IID (Identical and Independently distributed) and Non-IID data silos. For Non-IID scenarios, we assume $10$ shards of size $240$ assigned to each client. We run it over $10$ clients in total, selecting only a fraction of $0.5$ in each round for training. We add DP locally, that is, to each client model at every iteration and aggregate them to perform Federated Averaging. We observe the best accuracies with RoBERTa for the centralised DP implementation, particularly with $\epsilon=25$ with an accuracy of 62.6\%. BERT in a centralised DP setting does come close at $\epsilon=25$ with an accuracy of 60.03\%. The results also show that the accuracy decreases by adding FL to the DP implementations.

\begin{table}[t]
\centering
  \caption{Averaged Test Accuracies of FL and DPFL models}
  \resizebox{\columnwidth}{!}{%
  \begin{tabular}{|c|c|c|c|}
  \hline
  \textbf{Setup}&\textbf{Epsilon($\epsilon$)}&\textbf{BERT}	&\textbf{RoBERTa}\\
  \hline 
    \multirow{5}{4.5em}{Centralized DP}	
    &0.5 &{31.5$\pm$23.94} &{31.36 $\pm$ 26.35}\\
    & 5	&{37.48$\pm$20.42} &{38.34 $\pm$20.08}\\& 15 &{51.71 $\pm$14.71} &{51.34 $\pm$ 15.45}\\& 20 &{55.37 $\pm$ 5.49} &{55.54 $\pm$ 5.54}\\ 
    & 25 &{60.03 $\pm$ 1.37} &{62.6 $\pm$ 4.24}\\ 
    \hline
    \multirow{5}{4.5em}{DP-FL\\ IID}	
    &0.5 &{14.57 $\pm$ 2.86} &{20.11 $\pm$ 7.68}\\
    & 5	&{30 $\pm$25.6} &{30.04 $\pm$ 28.22}\\ 
    & 15 &{40.34 $\pm$ 20.55} &{50.26 $\pm$ 20.84}\\
    & 20 &{51.05 $\pm$ 7.95} &{54.78 $\pm$ 2.99}\\ 
    & 25 &{53.47 $\pm$ 6.48} &{61.38 $\pm$ 0.93}\\ 
    \hline
    \multirow{5}{4.5em}{DP-FL\\ Non IID}	
    &0.5 &{19.82 $\pm$ 5.97} &{33.13 $\pm$ 25.41}\\ 
    & 5 &{35.74 $\pm$ 21.48} &{36.51 $\pm$ 26.87}\\ 
    & 15 &{45.87 $\pm$ 15.56} &{49.83 $\pm$ 20.6}\\
    & 20 &{52.43 $\pm$ 4.08} &{53.36 $\pm$ 3.27}\\ 
    & 25 &{58.96 $\pm$ 2.56} &{60.83 $\pm$ 0.53}\\ 
    \hline
  \end{tabular}
  }
\end{table}

We also empirically observe that with increase in $\epsilon$ , accuracy of the models also increases. This happens because as the value of $\epsilon$ increases, privacy decreases with the addition of noise from a smaller range which results in smaller variance. Consequently, the accuracy of the model increases. Inherently, applying DP to deep learning yields loss of utility due to the addition of noise and clipping. We can also observe that the performance of federated language models still lies behind that of centralized ones.

\section{Conclusion}
Financial data is highly sensitive , hence the risks of collecting and sharing data can limit studies. Financial organizations work with a lot of confidential user data and therefore highly value protecting the data to retain the integrity of the user and we need to delve into research of private training of machine learning models to ensure this. During this study, we benchmark the utility of privacy models while attempting to preserve the performance of SOTA transformer models such as BERT and RoBERTa. Our empirical results show that the models show better performance with increasing  $\varepsilon$ as expected with the decrease in noise. The models come close to the performance of the baseline models near the higher $\varepsilon$ values.The DP + FL shows a similar trend which showcases a greater protection feature without compromising the performance. As future work, we hope to improve our models further by hyper-parameter tuning, freezing partial layers of the NLP model and implementing focal loss on the unbalanced dataset to better the results. The complete code to this paper can be found here:~\href{here}{https://www.github.com/tiasa2/Privacy-enabled-Financial-Text-Classification-using-Differential-Privacy-and-Federated-Learning}.

\bibliography{dp}
\bibliographystyle{acl_natbib}

\end{document}